\documentclass{article}

\usepackage{times}
\usepackage{soul}
\usepackage{url}
\usepackage{hyperref}
\usepackage[utf8]{inputenc}
\usepackage[small]{caption}
\usepackage{amsmath}
\usepackage{amsthm}
\usepackage{booktabs}
\usepackage{algorithm}
\usepackage{algorithmic}
\usepackage[switch]{lineno}
\usepackage{longtable}
\usepackage{supertabular}
\usepackage{multirow}

\usepackage{graphicx}
\usepackage{booktabs} %
\usepackage{makecell}
\usepackage{subfigure}

\usepackage{amsmath}
\usepackage{amssymb}
\usepackage{mathtools}
\usepackage{enumitem}

\usepackage{color}
\usepackage{diagbox}
\usepackage{caption}
\usepackage{wrapfig}

\usepackage[utf8]{inputenc} %
\usepackage[T1]{fontenc}    %
\usepackage{amsfonts}       %
\usepackage{nicefrac}       %
\usepackage{microtype}      %
\usepackage{xcolor}         %

\newcommand{\methodname}{LaMAI}

\def\gA{{\mathcal{A}}}

\def\gM{{\mathcal{M}}}

\def\gQ{{\mathcal{Q}}}

\def\gS{{\mathcal{S}}}

\usepackage[in]{fullpage}  
\usepackage{arxiv} 
\usepackage{mathpazo} 

\usepackage[utf8]{inputenc} 
\usepackage[T1]{fontenc}    
\usepackage{url}            
\usepackage{booktabs}       
\usepackage{amsfonts}       
\usepackage{microtype}      
\usepackage{lipsum}         

\usepackage{graphicx}  
\usepackage{float} 
\usepackage{subfigure} 
\usepackage{algorithm}
\usepackage{algorithmic}
\usepackage{tabulary}

\usepackage{amssymb}
\usepackage{pifont}

\hypersetup{
    colorlinks=true,
    citecolor=blue,
    linkcolor=blue,
    citecolor=orange
}

\title{Empowering Language Models with Active Inquiry for Deeper Understanding}

\usepackage{authblk}
\author{%
  Jing-Cheng Pang\textsuperscript{\rm 1,3,*}, 
  Heng-Bo Fan\textsuperscript{\rm 2,*}, 
  Pengyuan Wang\textsuperscript{\rm 1,3,*}, 
  Jia-Hao Xiao\textsuperscript{\rm 2,*}, 
  Nan Tang\textsuperscript{\rm 1,3}, 

  Si-Hang Yang\textsuperscript{\rm 1,3}, 
  Chengxing Jia\textsuperscript{\rm 1,3}, 
  Sheng-Jun Huang\textsuperscript{\rm 2,$\diamond$}
  Yang Yu\textsuperscript{\rm 1,3}\\
  \textsuperscript{\rm 1} National Key Laboratory for Novel Software Technology, Nanjing University, China \\
  \& School of Artificial Intelligence, Nanjing University, China \\
  \textsuperscript{\rm 2} College of Computer Science and Technology/Artificial Intelligence, \\ Nanjing University of Aeronautics and Astronautics, China \\
  \textsuperscript{\rm 3}Polixir.ai \\
  \textsuperscript{*} Equal contribution\\
  \textsuperscript{$\diamond$} Corresponding: huangsj@nuaa.edu.cn
}

\date{}
\begin{document}

\maketitle

\begin{abstract}
    The rise of large language models (LLMs) has revolutionized the way that we interact with artificial intelligence systems through natural language. However, LLMs often misinterpret user queries because of their uncertain intention, leading to less helpful responses. In natural human interactions, clarification is sought through targeted questioning to uncover obscure information. Thus, in this paper, we introduce \methodname~(Language Model with Active Inquiry), designed to endow LLMs with this same level of interactive engagement. \methodname~leverages active learning techniques to raise the most informative questions, fostering a dynamic bidirectional dialogue. This approach not only narrows the contextual gap but also refines the output of the LLMs, aligning it more closely with user expectations. Our empirical studies, across a variety of complex datasets where LLMs have limited conversational context, demonstrate the effectiveness of \methodname. The method improves answer accuracy from 31.9\% to 50.9\%, outperforming other leading question-answering frameworks. Moreover, in scenarios involving human participants, \methodname~consistently generates responses that are superior or comparable to baseline methods in more than 82\% of the cases. The applicability of \methodname~is further evidenced by its successful integration with various LLMs, highlighting its potential for the future of interactive language models.
\end{abstract}

\section{Introduction}

Recent advances in large language models (LLMs) \cite{gpt4,llama2,chatglm} lift the curtain of a new era in human-machine interaction. These language models, pre-trained on massive corpora of text \cite{Large_corpora}, are designed to interpret and respond to user queries with human-like proficiency. One of the most impressive applications of LLMs is the chatbot, which generates responses to various user queries and interacts with humans using natural language. However,  the inherent ambiguity and potential for misunderstanding in natural language present challenges \cite{talar}.
Users may unintentionally pose ambiguous queries or assume a shared context that LLMs do not access \cite{ask_cla_que_info_seek}. For example, when seeking advice on personal issues, users might assume the LLM knows their previous experiences, leading to a disconnect in the interaction.

Addressing ambiguous user queries remains a significant challenge for LLMs. The autoregressive nature of 
\begin{wrapfigure}[15]{rt}{0.5 \textwidth}
    \includegraphics[width=0.5 \textwidth]{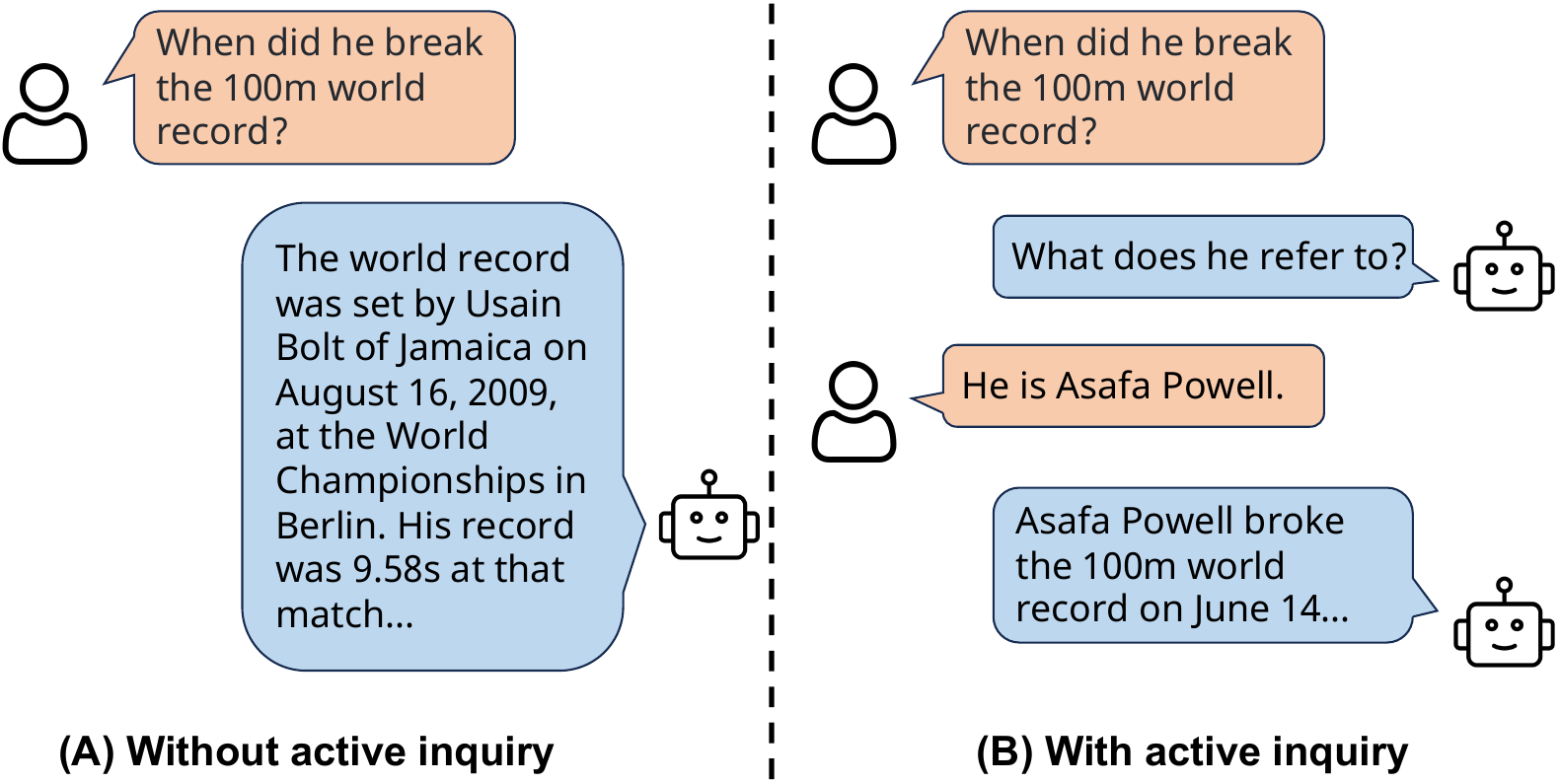}
    \vspace{-0.85em}
    \caption{
An illustrative example of language model active inquiry. (A) LLM directly answers the question without active inquiry. (B) LLM actively seeks clarification by inquiring about the user.}
\label{fig:comparison_rag_iag}
\end{wrapfigure}
text generation in LLMs often leads to direct response generation \cite{Hallucination20}, even when the query lacks context or specific information, which can result in unhelpful or incorrect responses. 
To tackle this issue, a prevalent strategy, Retrieval-Augmented Generation (RAG \cite{RAG}), adds information to LLMs by integrating them with external information sources, such as knowledge bases or search engines \cite{self_ask}. However, these methods primarily address the challenge of outdated knowledge within LLMs and do not adequately handle ambiguous user input \cite{kb_human_intent}.

Drawing inspiration from human communication, where clarification is sought through actively questioning, research has been conducted on developing AI systems that actively asks users \emph{clarifying questions}. This strategy has been successfully applied in certain scenarios such as hotel reservation \cite{hotel_res_system} and food ordering \cite{food_order}.
Nonetheless, there is still a gap in equipping LLMs with this capability to actively seek information from users and deal with ambiguous user queries.
To bridge this gap, we introduce the \underline{La}nguage \underline{M}odel \underline{A}ctive \underline{I}nquiry (\methodname) method, which actively inquires the user by posing targeted clarifying questions. Fig.\ref{fig:comparison_rag_iag} presents an illustrative example of \methodname. Specfically, before answering the user query, \methodname~estimates the LLM's uncertainty regarding the query by sampling multiple responses and calculating their variation. Under high uncertainty, LLMs needs to actively inquire the user for clarification. To achieve this, \methodname~prompts the model to generate a series of questions about the user query. Besides, \methodname~employs active learning to select and formulate the most informative queries, thus engaging in a dynamic information exchange. By doing so, \methodname~is able to obtain the missing pieces of context directly from the user. The process of active inquiry not only refines the interaction but also significantly enhances the quality and relevance of the LLM's responses.

The main contributions of this work are as follows:
Firstly, we introduce a new answer generation scheme for LLMs, which emphasizes active information acquisition through inquiring the user. This generation process enables LLMs to obtain precise and query-specific information, distinguishing from traditional methods that rely on static knowledge bases.
Secondly, we propose a novel method, \methodname, which involves active learning techniques to enable effective and efficient interaction with the user.
Thirdly, we conduct comprehensive experiments across various tasks and demonstrate that \methodname~consistently outperforms existing question-answering frameworks on various LLMs. These results highlight the robustness of the \methodname~approach, demonstrating its efficacy in improving LLM's comprehension of user input, and thus enhancing the quality of language model response.

\section{Related Work}
\label{sec:related_work}

\subsection{Enhancing User Input Understanding}
Previous studies have explored the concept of constructing AI systems that engage users by posing questions for clarification or information \cite{ask_cla_que_info_seek,ask_clarifying_questions_web_search,ask_cla_dialog}. 
Such systems have been effectively implemented in domains like hotel reservation services \cite{hotel_res_system}, where they prompt users with specific inquiries to verify and complete booking details. The field of natural language processing has also shown an interest in generating clarifying questions, particularly in response to ambiguous queries.
For example, \cite{ask_clarifying_questions_web_search} utilized a set of question templates to address the issue of ambiguous web search queries. \cite{ask_cla_dialog} studied the task of selecting clarifying questions from a set of human-generated questions for open-domain information seeking. A separate investigation \cite{ask_clar_entity} concentrated on the task of asking clarifying questions for entity disambiguation, phrased as "Did you mean A or B?" However, this method is limited to entity disambiguation and does not apply to a broader range of queries, such as those with multiple facets. Though various research studies seek clarification from users, there has been a gap in equipping LLMs with this capability. This paper introduces the \methodname~method, which enhances LLMs' understanding of user input, improving their performance.

\subsection{Improving Language Model Response}

Research on improving LLM response has attracted considerable attention \cite{ask_clarifying_questions_web_search}. Related methods can be divided into two categories: invasive methods that train LLM to generate high-quality response \cite{SIRLC}, and non-invasive methods that improve generation by simply providing demonstrations of the task \cite{brown2020language}, or increasing model's reasoning ability through various answer refinement techniques \cite{CoT,self_consistency}. However, a huge limitation in these prompting techniques is that they cannot elicit knowledge absent from training data, which leads to hallucinated response \cite{huang2023survey}, especially in specialized domains \cite{gpt4}. 
A related line of approaches improves model response by allowing the model to access external knowledge or information. These proposed methods, known as Retrieval-Augmented Generation (RAG) 
\cite{IzacardLLHPSDJRG23,ALLMSurvey23,JiangXGSLDYCN23}, aim to incorporate LLM with external knowledge to enrich its final response. A retriever is commonly used for searching related content based on keywords \cite{RAG} and is jointly fine-tuned with a sequence-to-sequence model.
Despite their potential, RAG-based methods primarily address the challenges of outdated, general knowledge rather than resolving specific, ambiguous user queries.
In our work, we propose a novel technique, \methodname, which empowers LLMs to seek clarifications from users actively. This approach enables LLM to handle ambiguities by inquiring additional information, thereby generating more accurate responses.

\subsection{Active Learning}
Active learning \cite{AL_inform} is a machine-learning approach that designs an effective sampling strategy to select the most valuable examples and query an oracle for their labels, to maximize model performance while minimizing annotation cost. 
Most of the existing approaches can be divided into two categories: the informativeness-based methods and the representativeness-based methods.
The former group primarily samples examples close to the decision boundary to reduce model uncertainty \cite{you2014diverse,yan2018cost}. \cite{huang2016active,zhang2017active} explored a variation for neural networks using gradient information as the metric of informativeness in text classification tasks. At the same time, the representativeness-based methods constrain the chosen data points to be distinct from each other or conform them to the data distribution \cite{roy2001toward,li2020deep}. N-gram or word counts can be regarded as a measure of density distribution and standards for sampling practical examples \cite{ambati2012active,zhang2021cartography}. 
Preference for instances with more unseen n-grams \cite{erdmann2019practical} is another approach to selecting more representative samples.
In this work, \methodname~utilizes active learning techniques to select the most informative questions for the user.

\begin{figure*}[t]
    \centering
    \includegraphics[width=0.95\linewidth]{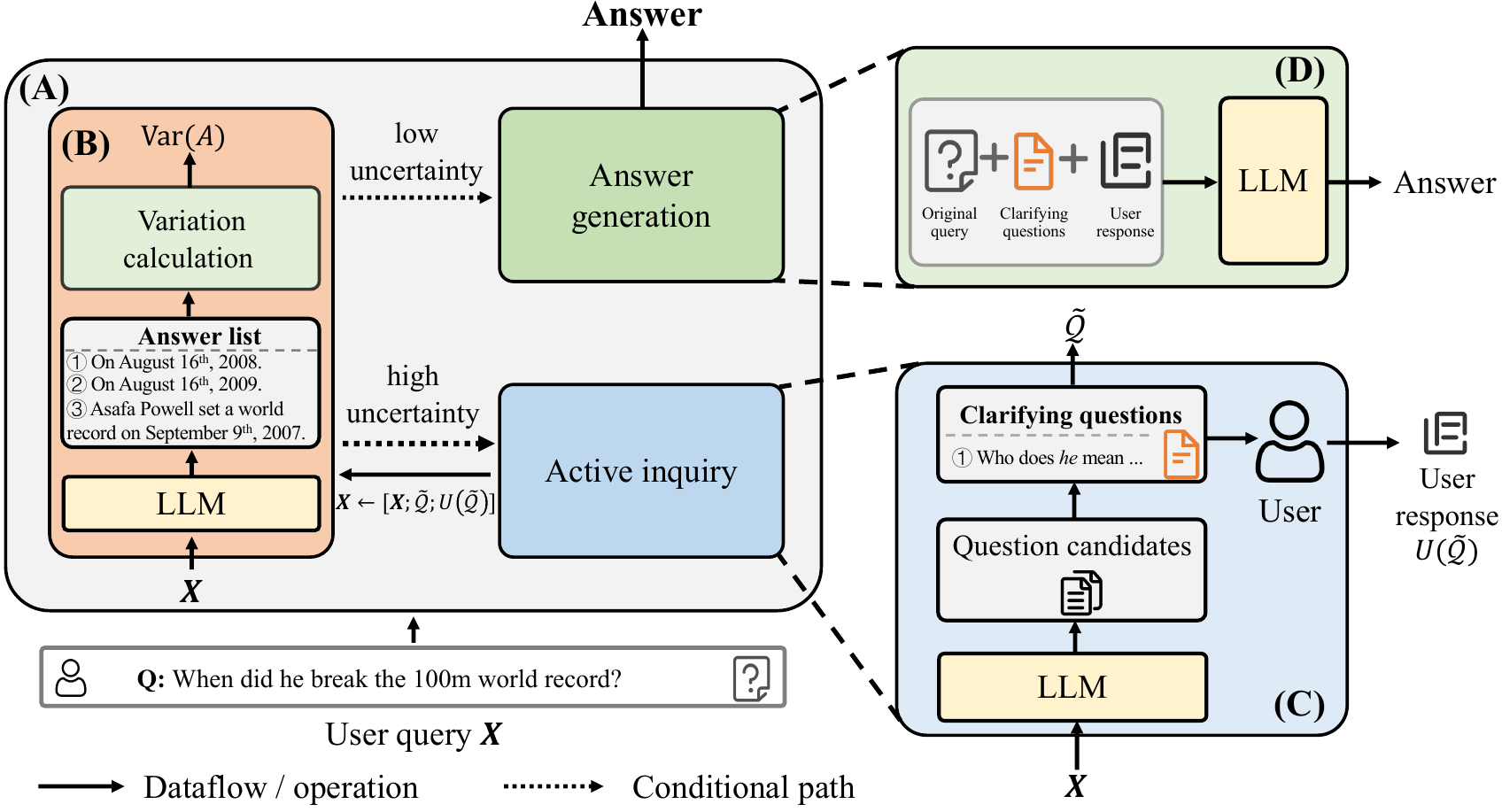}
    \caption{
    Illustration of \methodname~method. \textbf{(A)} Overall workflow: the user query $\boldsymbol{X}$ is processed by the uncertainty estimation module. Depending on the outcome, LLM either inquires the user with clarifying questions or directly generates the response. \textbf{(B)} Uncertainty estimation module evaluates the LLM's level of uncertainty regarding a query $\boldsymbol{X}$. \textbf{(C)} In the active inquiry module, LLM actively inquires the user with clarifying questions. After receiving the user's feedback, \methodname~updates the user query to incorporate this new information and re-estimates the uncertainty. \textbf{(D)} Answer generation module generates the answer to a user query. 
    }
    \label{fig:overall_framework}
\end{figure*}

\section{Method}

This section introduces our primary contribution, Language Model Active Inquiry (\methodname), which equips LLM with the ability to actively inquire the user for a better understanding of the user intent. We describe the problem formulation (Sec.\ref{sec:prob_formulation}), the details of \methodname~(Sec.\ref{sec:method_workflow}), and practical implementation of \methodname~(Sec.\ref{sec:prac_imp}).

\subsection{Problem Formulation}
\label{sec:prob_formulation}

We consider a typical scenario where the user is having a conversation with the LLM $\mathcal{M}$, which takes the user query $\boldsymbol{X}$ as input and outputs the answer based on the user query: $\boldsymbol{Y}=\mathcal{M}(\boldsymbol{X})$. We expect LLM to generate a helpful response to the user query.

However, the user query often unintentionally leaves out some critical context/information that LLM does not access, resulting in the LLM's misunderstanding of such ambiguous user query. To seek clarification from the user, LLM actively inquires the user with a set of clarifying questions $\Tilde{\gQ}$. Then, the user will provide the feedback  $U(\Tilde{\gQ})$ to the questions as a supplemental clarification to the initial user query. The interaction above can be repeated several times to ensure that the LLM accurately grasps the intent of the user. Finally, the LLM outputs the answer based on the original user query and the user feedback.

\subsection{Language Model Active Inquiry}
\label{sec:method_workflow}

This section introduces our proposed \methodname~method, with the overall workflow presented in Fig.\ref{fig:overall_framework}. Given a user query $\boldsymbol{X}$, \methodname~evaluates LLM's uncertainty regarding the query. If the uncertainty is high, \methodname~inquires the user to clarify the ambiguity in the initial user query and then augments the user query with user feedback as additional information. Then \methodname~re-estimates the uncertainty regarding the updated query. This process is repeated until the LLM's uncertainty about user query is satisfactorily reduced.
\methodname~consists of three key components: (1) \emph{uncertainty estimation} that evaluates LLM's uncertainty about a user query $\boldsymbol{X}$, (2) \emph{active interaction} that asks user clarifying questions to better understand user's intent, and (3) \emph{answer generation} that generates the answer. We will elaborate on these three components in the following.

\subsubsection{Uncertainty Estimation by Multiple Answers Sampling}
The uncertainty estimation module evaluates LLM's uncertainty about a user query $\boldsymbol{X}$ and determines when to actively inquire about the user. The previous study verifies that the uncertainty estimation can be derived from an ensemble of generated answers \cite{uncertainty_estimation}. To achieve this, \methodname~samples multiple answers $\{A_1, A_2, \cdots, A_{\rm T} \}$, where each answer $A_i=\gM(\boldsymbol{X})$ is generated by the LLM under appropriate temperature parameters. To estimate the uncertainty, \methodname~applies an embedding model to convert the array of answers into corresponding text embeddings $\{E_1, E_2, \cdots, E_{\rm T} \}$. A high variance in these embeddings indicates greater uncertainty by the LLM regarding the query, whereas a low variance suggests confidence. The variation of the answer is calculated as follows:
\begin{equation}
  \label{eqvariance}
  \operatorname{Var}(A) = \frac{1}{K} \sum_{k=1}^{K} \left( \frac{1}{T-1} \sum_{i=1}^{T} (E_{i}^k - \bar{E}^k)^2 \right),
\end{equation}
where $K$ is the dimensionality of the embeddings, $T$ is the number of sampled responses, $E_{i}^k$ denotes the value of the $i$-th answer for the $k$-th dimension, and $\bar{E}^k = \frac{1}{T} \sum_{i=1}^{T} E_{i}^k$ is the mean value for the $k$-th dimension across all $T$ embeddings.

Subsequently, \methodname~operates on distinct pathways, conditioned on the uncertainty estimation outcome.
If the uncertainty is low, i.e., $\operatorname{Var}(A) \leq \delta$ ($\delta$ denotes the active inquiry threshold), \methodname~outputs the answer $A=\gM(\boldsymbol{X})$. Otherwise, in cases of higher uncertainty, it indicates that LLM requires more clarification from the user for a better understanding of the user query.

\subsubsection{Active Inquiry with Selective Clarifying Questions}

Active Inquiry Module (AIM) enables LLM to interact with the user, ask clarifying questions, and better understand the user's intent. The primary focus is on how to formulate practical clarifying questions. A direct yet effective approach is leveraging LLM to generate potential questions \cite{llm_ask_questions}. Specifically, AIM prompts the LLM to generate a set of questions $\gQ$ regarding the user's input $\boldsymbol{X}$. The specific prompts used in the experiments are presented in Appendix \ref{sec:app_exp_details}.
Nevertheless, questions generated in this manner may not be uniformly helpful and could increase the user's burden by raising too many questions simultaneously. To mitigate this, filtering and presenting only the most helpful and insightful questions to the user is essential.
Active learning, recognized for its proficiency in identifying informative and representative samples \cite{AL_inform}, can be naturally applied to selecting clarifying questions. Given a set of $N$ potential questions $\gQ$, the goal is to select a subset of $M$ ($M\leq N$) clarifying questions $\Tilde{\gQ}=\gS(\gQ)$. For this selection process, we explore two prevalent active learning strategies, denoted as $\gS$.

\begin{enumerate}
    \item \textbf{Similarity} strategy focuses on identifying questions that closely align with the user query. To achieve this, \methodname~first extracts all questions and user query embeddings. Cosine similarity is then employed to measure the degree of semantic correspondence among embeddings, allowing for selecting the top $M$ questions that exhibit the greatest relevance to the user's query. The underlying rationale behind this strategy is the intuition that a higher degree of similarity correlates with a richer provision of pertinent information to address the user's query.
    \item \textbf{Diversity} strategy aims to capture a broad spectrum of questions. This is accomplished by encoding all questions within the set $\gQ$ using an embedding model. Following this, the K-Means clustering algorithm \cite{KMeans1967} is applied to categorize the question embeddings into $M$ clusters. From each cluster, one question is chosen at random. This method ensures that the selected questions represent a diverse and extensive range of information, which can enhance the comprehensiveness of the questions gathered.
\end{enumerate}

The system presents the user with this set of clarifying questions $\Tilde{\gQ}$. The user responds to these questions successively, providing clarification, denoted as $U(\Tilde{\gQ})$. 
Following this, the original user query is augmented to include the user's clarifications, resulting in an updated query: $\boldsymbol{X}\leftarrow [\boldsymbol{X};\Tilde{\gQ},U(\Tilde{\gQ})]$, which we call it \emph{interaction-augmented query}. This augmentation draws upon the directional stimulus prompt technique \cite{direct_sp}, which has been shown to improve the model generation. 
Subsequently, \methodname~re-estimates LLM's uncertainty regarding the updated user query. This uncertainty estimation and active inquiry process continues iteratively until the uncertainty falls below the active inquiry threshold $\delta$ or the maximum number of iterations is reached.

\subsubsection{Answer Generation with Interaction-augmented Query}
With the interaction-augmented query induced by AIM, the LLM can generate the answer to this query enriched with the additional context of user clarifications. Note that \methodname~does not necessarily inquire the user if the LLM's uncertainty about the initial user query has already been low. For clarity, the answer generation process is described here and in Fig.\ref{fig:overall_framework}(D) under the condition that the user interaction occurs.

\subsection{Practical Implementation}
\label{sec:prac_imp}

In the design of \methodname, we present an iterative interaction between the user and the LLM. While this iterative process can be laborious and time-consuming due to the requirement for the user to respond to clarifying questions, our practical implementation streamlines this to a single user interaction iteration. This refinement is sufficient to elucidate the user's initial query. We present this single-iteration implementation process in Algorithm \ref{AICL_algo}, detailed in Appendix \ref{sec:algo}. We apply \texttt{text-embedding-ada-002} model \cite{text_embedding} to convert text to embeddings. 
\section{Experiment}
\label{sec:exp}

In this section, we evaluate the efficacy of \methodname~through extensive experiments on diverse, challenging datasets.
The objective is to validate \methodname's capacity to improve LLM's understanding of user queries and to improve the quality of the responses provided.
We aim to answer the following important questions: (1) How does \methodname~perform in comparison to current answer generation methods across different themes of user query (Sec.\ref{sec:exp_main},\ref{sec:exp_human_invol})? (2) How does \methodname~improve the comprehension ability of LLM regarding user queries (Sec.\ref{sec:exp_main})? (3) Can \methodname~be integrated with various sizes of LLMs (Sec.\ref{sec:exp_on_diff_llm})? and (4) What is the impact of each component and parameter on the performance of \methodname~(Sec.\ref{sec:exp_ablation})? We begin with introducing the experimental setting.

\subsection{Experimental Setting}
In this subsection, we introduce the datasets used in the experiments, followed by the introduction of the baseline methods. Finally, we present implementation details and evaluation settings in our experiments.

\textbf{Dataset.} We conduct experiments on five challenging Q\&A datasets alongside a dataset dedicated to meeting summarization. Each problem in these datasets comprises a user query paired with supporting facts that serve as a context for the query. This property makes these datasets a perfect testbed for evaluating our  \methodname~framework where user-input-oriented context is missed. Tab.\ref{tab:example_of_dataset_main_body} shows some examples from these datasets. In our experiments, all methods, if not specifically marked, are agnostic to the supporting facts. These datasets are: 
(1) \textbf{HotpotQA} \cite{hotpotqa} is collected on the English Wikipedia. Each question in the dataset comes with the two gold paragraphs, as well as a list of sentences that crowdworkers identify as supporting facts necessary to answer the question;
(2) \textbf{StrategyQA} \cite{strategyqa} consists of questions similar to that in HotpotQA, but the answer format is limited to True or False;
(3) \textbf{2WikiMultiHopQA} \cite{2wiki} uses structured and unstructured data and introduces the evidence information containing a reasoning path for multi-hop questions; 
(4) \textbf{MuSiQue} \cite{MuSiQue} is a multi-hop QA dataset with 2-4 hop questions using seed questions from five different single-hop datasets; 
(5) \textbf{IIRC} \cite{iirc} is a dataset for incomplete information reading comprehension, providing only partial information to answer them, with the missing information occurring in one or more linked documents;
(6) \textbf{QMSum} \cite{QMSum} contains dialogue histories of multi-domain meeting. The user queries LLM to answer specific questions about the meeting, while some parts of the dialog are masked out.

\begin{table}[htbp]

    \centering
    \begin{tabular}{c|p{2.4cm}|p{3cm}}
   \toprule
    & HotpotQA & StrategyQA \\
   \midrule
   User query & Were Up and The Watercolor released in the same year? & Are more people today related to Genghis Khan than Julius Caesar?  \\ 
   \midrule
   \makecell[tc]{Supporting \\ facts} & Up and The Watercolor are two films. Up was released in ... & Compare the number of their offspring. Julius Caesar had three children. Genghis Khan had sixteen children ... \\ \midrule
   Label & Yes & True \\  
   \bottomrule
\end{tabular}

    \caption{Examples of user query, supporting facts, and correct answer for the datasets used in the experiments. }
    \label{tab:example_of_dataset_main_body}
\end{table}

\begin{table*}[t]
\scriptsize
    \centering
    \begin{tabular}{c|ccc|ccc|ccc|ccc|ccc}
    \toprule
   \diagbox{Method}{Dataset}& \multicolumn{3}{c|}{HotpotQA}  & \multicolumn{3}{c|}{StrategyQA} & \multicolumn{3}{c|}{2WikiMultiHopQA} & \multicolumn{3}{c|}{MuSiQue} & \multicolumn{3}{c}{IIRC}  \\
   \midrule
                        & EM & F1 & Acc & EM & F1 & Acc & EM & F1  & Acc & EM & F1  & Acc & EM & F1  & Acc \\
   DG                   & 29.1 & 38.3 & 44.4 & 57.2 & 57.6 & 57.6 & 17.3 & 22.9 & 43.3 & 3.8 & 14.0 & 20.5 & 14.7 & 18.1 & 20.8  \\
   CoT                  & 32.6 & 41.6 & 48.1 & 66.7 & 66.9 & 66.9 & 31.0 & 32.9 & 49.6 & 5.5 & 12.8 & 19.8 & 17.5 & 22.0 & 24.6  \\
   Self-ask             & 27.7 & 38.0 & 58.6 & 34.3 & 63.2 & 34.3 & 42.5 & 49.2 & 54.0 & 15.0 & 27.0 & 28.5 & 10.1 & 30.1&16.8  \\
   RAG (web)            & 16.2 & 25.4 & 47.9 & 33.1 & 63.3 & 33.1 & 28.7 & 36.9 & 51.1 & 10.0 & 20.2 & 27.8 & 6.0 & 10.7 & 25.6  \\
   \midrule
   \methodname          & 47.5 & \textbf{58.7} & \textbf{68.1} & 66.3 & 66.4 & 66.4 & 42.8 & 52.0 & \textbf{71.3} & 16.5 & 25.9 & \textbf{30.5} & \textbf{34.1} & \textbf{42.6} & \textbf{51.1} \\
   \methodname+CoT      & \textbf{49.1} & \textbf{59.2} &	\textbf{69.6} & \textbf{71.7} & \textbf{71.7} & \textbf{71.7} &	\textbf{49.8} & \textbf{61.1} & \textbf{73.0} & \textbf{18.5} & \textbf{27.7} & \textbf{31.5} & 27.2 & 36.6 & 45.1   \\
   \midrule
   Oracle & 59.4 & 86.9 & 72.4 & 78.2 & 78.2 & 78.1 &  60.5 &	72.0 & 83.5	 & 21.0	& 33.5	& 36.0 & 29.4 & 62.5 & 42.6 \\
   \bottomrule
\end{tabular}
    \caption{Comparative experimental results on Q\&A datasets. \methodname~surpasses oracle method that takes supporting facts as input on certain datasets. Acc stands for the accuracy evaluated by ChatGPT. Excluding Oracle's results, the best (or nearly best) results are highlighted in \textbf{bold}. 
}
    \label{tab:main_results}
\end{table*}

\textbf{Baseline.} 
We compare \methodname~with five representative baseline methods that are widely applied in the community. 
(1) Direct Generation (\textbf{DG}) directly generates the answer with the deterministic output of the LLM;
(2) Chain of Thought (\textbf{CoT}) \cite{CoT} generates the answer by prompting the LLM to reason the results through a series of intermediate reasoning steps. We use the `\emph{Let's think step by step}' prompt in the experiments;
(3) \textbf{Self-ask} \cite{self_ask} poses and responds to the self-generated follow-up questions, refining its understanding before providing the initial query's response;
(4) \textbf{RAG (Web)} extends the Self-ask method by integrating a web-search API, specifically the Bing search\footnote{Bing search API is accessible at \url{https://www.microsoft.com/en-us/bing/apis/bing-web-search-api}.}, to incorporate external knowledge into its answering process;
(5) \textbf{Oracle} generates the answer directly but is provided ground-truth supporting facts.

\textbf{Evaluation and Metrics.}
We evaluate the first 400 questions of the training set across five Q\&A datasets.
We utilize the in-context learning approach for all methods, selecting two random examples from each dataset to serve as demonstrations. This strategy ensures that the language model generates answers in a standardized format. We evaluate performance using exact match (EM) and F1 scores across the Q\&A datasets.
Following \cite{shao2023enhancing}, we also evaluate answer accuracy (Acc) using the \texttt{gpt-3.5-turbo} model, given that direct calculation of answer accuracy is not feasible on these datasets.
For the QMSum dataset, which contains long text, we apply \texttt{gpt-4} to evaluate the preference over the DG method.

\textbf{Implementation Details.}
We utilize ChatGPT \cite{gpt4} as the primary model for our experiments, a popular model in the field. To reduce the human costs in providing feedback to clarifying questions, we deploy ChatGPT in dual capacities: as an LLM (specifically, GPT-3.5) and as a pseudo-human interlocutor (GPT-4) within our experimental framework\footnote{The experiments are conducted between December 1, 2023, and January 15, 2024, employing the \texttt{gpt-3.5-turbo} and \texttt{gpt-4} models.}. The experimental design presents queries from the dataset to the LLM, while the supporting facts are exclusively provided to the GPT-4 model. In this way, the user query lacks the necessary context or information to be answered. We also consider \methodname+CoT method, which equips \methodname~with \emph{Let's think step by step} prompt.
More detailed information regarding the experimental setting, such as prompts and parameters, is available in Appendix \ref{sec:app_exp_details}.

\subsection{Main Results}

\label{sec:exp_main}

\textbf{Comparison with Baselines on Q\&A datasets.} 
Tab.\ref{tab:main_results} presents the comparative performance of \methodname~against baseline methods across five Q\&A datasets.
In general, \methodname~consistently surpasses the baseline methods, achieving an average answer accuracy of 50.9\%, which marks a significant improvement compared to the traditional generation method DG (31.9\%). This result underscores \methodname's superiority in generating helpful responses. 
The coT method, specifically designed to enhance the reasoning capabilities of LLM, also falls short of \methodname's performance. The limitation of CoT for our setting lies in its reliance on the LLM's embedded knowledge, which does not actively seek out additional context that may be absent in the user's query.
Similarly, the Self-ask strategy, which also leverages the LLM's embedded knowledge, cannot address the gaps in information presented by the user, resulting in a performance disparity with \methodname.
RAG (web) is a typical RAG-based method that seeks information from the web. The experiment results of RAG (web) suggest that the external knowledge base does not demonstrate an advantage in dealing with ambiguous user queries, as indicated by its close performance to the DG method. Notably, \methodname~achieves comparable performance with the Oracle method on HotpotQA and StrategyQA and outperforms it on IIRC. These results provide strong evidence that \methodname~
effectively acquires valuable information from the user, improves the model's comprehension of user queries, and leads to improved performance.

\begin{figure*}[ht]
\centering

\subfigure[GPT-4 Evaluation]{
\label{fig:gpt4_eval}
    \includegraphics[width=0.31\linewidth]{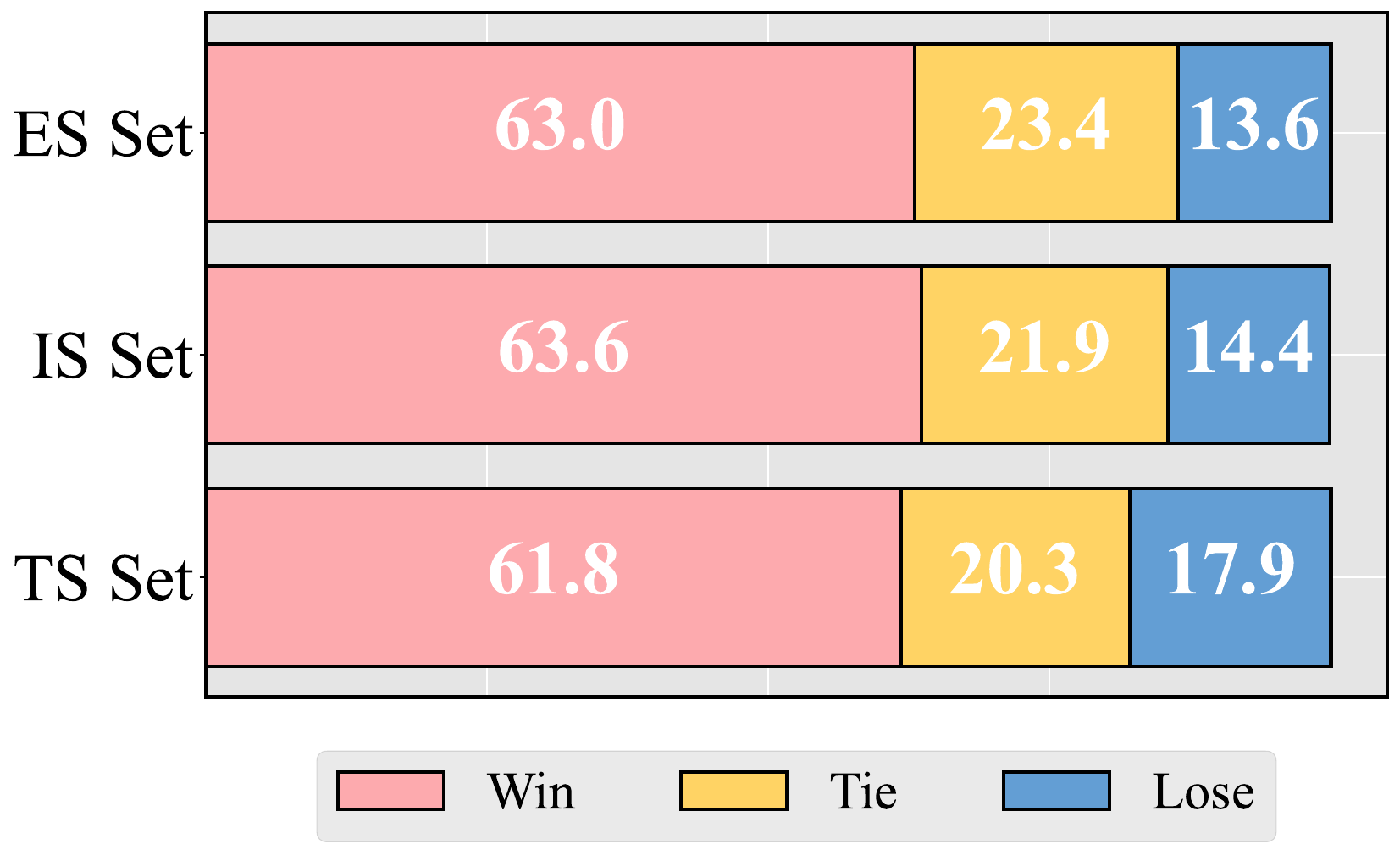}
}
\subfigure[Human-participated Experiment]{
\label{fig:human_eval}
    \includegraphics[width=0.31\linewidth]{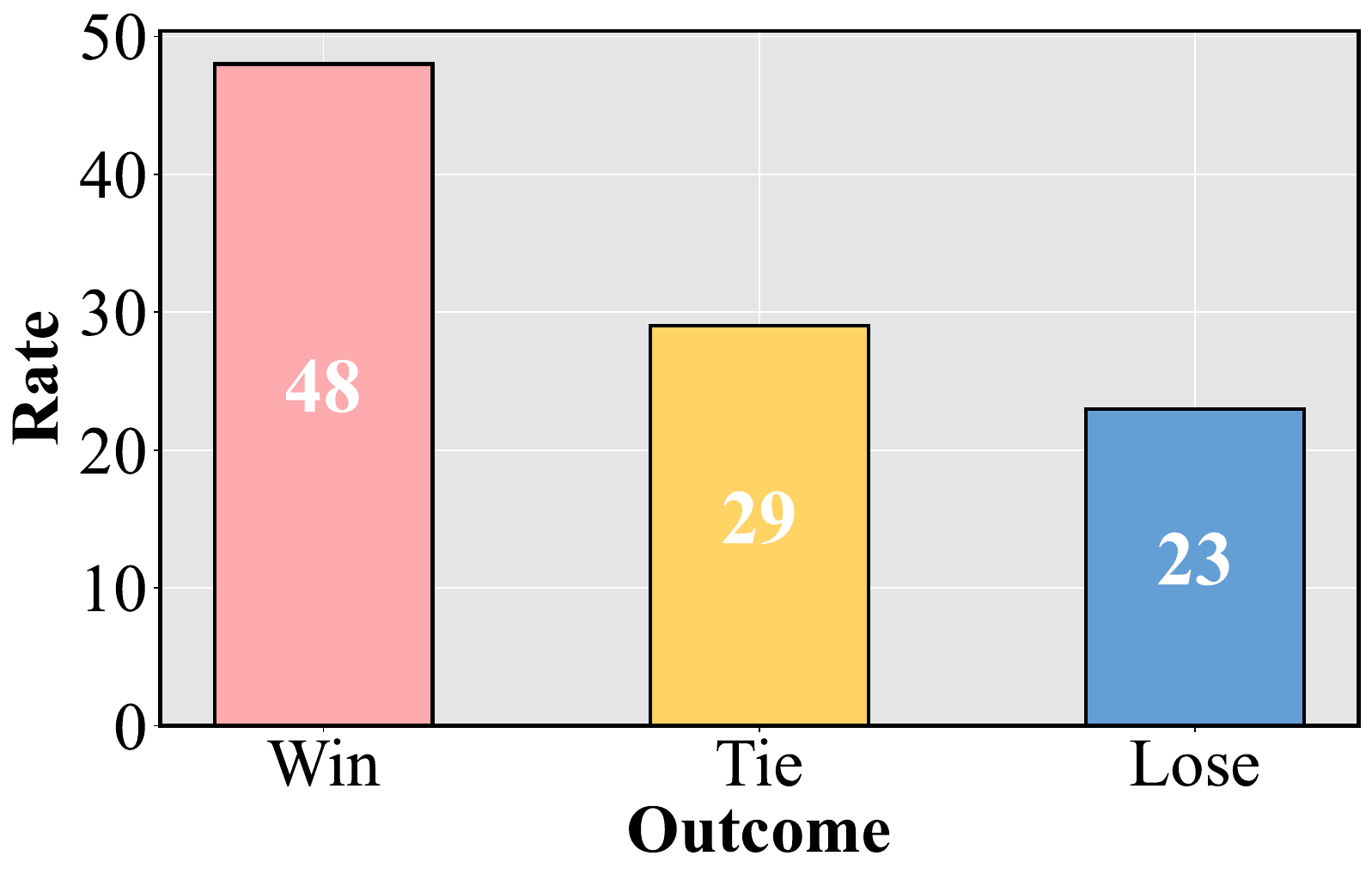}
}
\subfigure[Ablation Study]{
\label{fig:ablation}
    \includegraphics[width=0.31\linewidth]{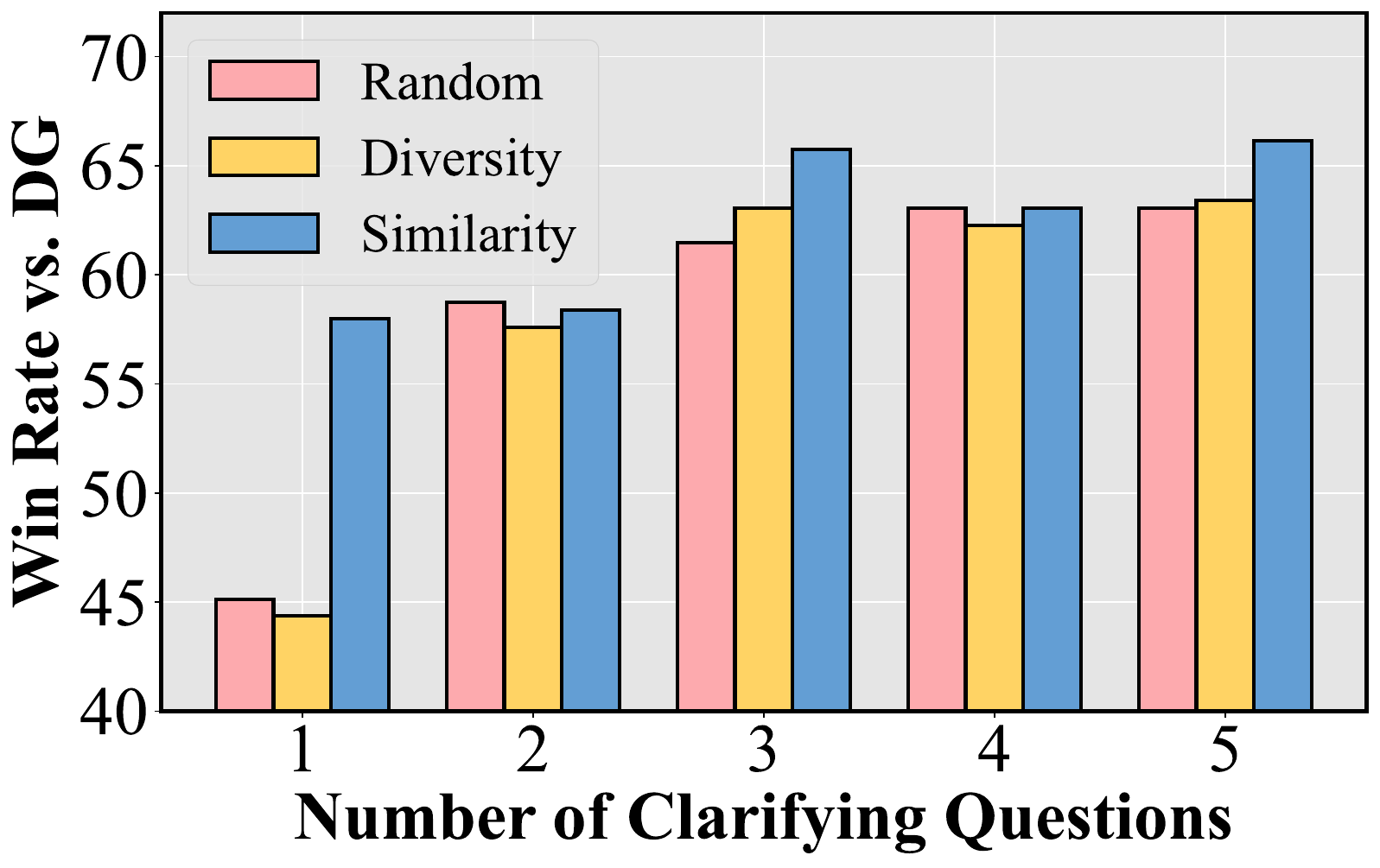}
}
\caption{
Comparative and analytical results for \methodname~on the QMSum dataset: 
\textbf{(a)} GPT-4-based evaluation showing \methodname's win rate against DG on three subsets of QMSum.
\textbf{(b)} Human-evaluated win rate of \methodname~against DG. In this experiment, \methodname~interacts with a human participant.
\textbf{(c)} Ablation study on the impact of the number of clarifying questions and active learning strategies on ES dataset. The values in the figure represent the win rate of \methodname~against DG method. 
}
\label{expfig:diff_hint_num}
\end{figure*}

\textbf{GPT-4 evaluation on QMSum.} 
We conduct experiments on the QMSum dataset, wherein the language model is asked to summarize specific perspectives based on the meeting's dialogue history. Assessing the quality of the summaries generated by the model is a significant challenge, as traditional reference-based metrics like BLEU and ROUGE often correlate poorly with human judgment. In our study, we employ GPT-4 as a reference-free metric for evaluation, which has demonstrated a high degree of alignment with human assessments \cite{gpt4_evaluation}.
Fig.\ref{fig:gpt4_eval} illustrates that \methodname~surpasses the baseline method, designated as DG, by a margin exceeding 60\% in win rate and 20\% in tie rate. These results underscore the efficacy of \methodname, which, through active interaction, garners valuable information from users to accurately respond to queries, thereby demonstrating the effectiveness of our approach.

\label{sec:case_study}

\textbf{Analysis on model outputs.} Previous results have demonstrated \methodname~enhances the capability of LLM to generate improved responses. To investigate the origins of this enhancement and its impact on the LLM's comprehension of user queries, we conduct a case study analyzing the model's outputs, which includes successful case and failed case, as shown in Tab.\ref{tab:example_of_model_output}. Our analysis yields several critical insights.
Firstly, \methodname~can raise practical clarifying questions to seek clarification for the user query. For instance, $\Tilde{\gQ}_1$ in the success case, \methodname~inquires about the specific actions that lead to injuries, aligning precisely with the supporting facts.
Secondly, applying active learning strategies filters the most informative questions from the question candidates.
Despite \methodname's generally superior performance, it occasionally generates questions that do not reflect the information in the supporting facts, such as $\Tilde{\gQ}_1$ in the failure case. This highlights an area for potential enhancement, specifically in the model's ability to generate relevant clarifying questions consistently. Addressing this could be a promising avenue for future research.

\begin{table*}[htbp]
\footnotesize
    \centering
\begin{tabular}{c|p{6cm}|p{6cm}}
   \toprule
   User query & \emph{Success case}: Is waltz less injurious than slam dance? & \emph{Failure case}: Can Billie Eilish afford a Porsche? \\ 
   \midrule
   \makecell*[tc]{Supporting \\ facts} &The waltz is a rhythmic dance performed in triple time by a couple. A slam dance is a type of dance in which leaping dancers collide against each other. & Billie Eilish is a famous female singer. Billie Eilish is 18 years old and has a net worth of \$25 Million. A Porsche Boxster is a car that starts at \$59,000. \$25,000,000 is greater than \$59,000.\\
   \midrule
   \makecell*[tc]{Questions before \\ selection} &  $\gQ_1$: What are the potential injuries associated with slam dancing? $\gQ_2$: Are there any specific movements or techniques in waltz that could lead to injuries? ... (10 questions in total) &  $\gQ_1$: Does Billie Eilish have any other expensive assets or investments? $\gQ_2$: Does Billie Eilish have any endorsement deals or sponsorships that could contribute to her ability to afford a Porsche? ...(10 questions in total) \\ \midrule
   \makecell*[tc]{Questions after \\ selection} & $\Tilde{\gQ}_1$: Are there any specific movements or techniques in waltz that could lead to injuries? $\Tilde{\gQ}_2$: Are there any specific movements or techniques in slam dancing that could lead to injuries?
   & $\Tilde{\gQ}_1$: Are there any known sources of income for Billie Eilish besides her music career? $\Tilde{\gQ}_2$: What is the average price range for a Porsche? 
   \\ \midrule
   Output / Label & True / True & False / True \\ 
   \bottomrule
\end{tabular}
    \caption{A case study on the model output for \methodname, with a success and a failure cases. We present more examples in Appendix \ref{sec:app_exp}.}
    \label{tab:example_of_model_output}
\end{table*}

\subsection{Human-in-loop Evaluation}
\label{sec:exp_human_invol}

In addition to experiments employing GPT-4 as a simulated human interlocutor, we also conduct human-participated experiments where actual human feedback is integrated. The experiment is based on the QMSum dataset, from which 100 user queries are randomly selected for evaluation. We invite five participants to the experiment, each accounting for 20 queries. Participants respond to clarifying questions posed by the LLM, which subsequently generates answers informed by the human response. Subsequently, another participant compares the answers generated by \methodname~and DG and annotates a preference for one over the other.
The results of the experiment are illustrated in Fig.\ref{fig:human_eval}. The human evaluation results reveal that \methodname~surpasses DG in 48\% of the instances, demonstrating its superior ability to comprehend user input during human interaction. Conversely, DG's responses are preferred in only 23\% of the cases, underscoring the effectiveness of \methodname~in the actual application scenario.

\subsection{Applicability on Different LLMs}
\label{sec:exp_on_diff_llm}

We verify the applicability of \methodname~by assessing its performance on different LLMs. We conduct experiments on 2WikiMultiHopQA and MuSiQue with Vicuna-13B \cite{vicuna}, a popular open-sourced LLM. The experimental setup is consistent with the main experiments, and the results are shown in Tab.\ref{tab:diff_llm}. The results demonstrate that \methodname~surpasses the DG and CoT methods on two datasets. This suggests that \methodname~is well-suited to models with fewer parameters. It is noteworthy that \methodname~exceeds or is comparable to \methodname+CoT method on the Vicuna-13B model. This could be attributed to the previous observation that CoT tends to perform more effectively with larger-sized LLMs \cite{flan-t5}.

\begin{table}[htbp]
    \centering
    \begin{tabular}{c|ccc|ccc}
    \toprule
   & \multicolumn{3}{c|}{2WikiMultiHopQA} & \multicolumn{3}{c}{MuSiQue}  \\
   \midrule
                        & EM & F1 & Acc & EM & F1 & Acc \\
    DG & 10.0 &	18.2 &	32.3 &	9.0 & 8.1 &	15.3 \\
    CoT & 30.8 & 35.8 &	40.5 & 11.5 & 12.1 & 17.2 \\
    \midrule
\methodname & 12.3 & 24.1 &	58.5 &	10.0 &	13.5 &	\textbf{33.0} \\
\methodname+CoT & \textbf{32.5} & \textbf{41.1} &	\textbf{63.3} &	\textbf{12.0} &	\textbf{18.8} &	30.2 \\
   \bottomrule
\end{tabular}
    \caption{Results on two datasets with Vicuna-13B.}
    \label{tab:diff_llm}
\end{table}

\subsection{Ablation Study}
\label{sec:exp_ablation}

In this subsection, we conduct an ablation study to investigate the influence of different components of \methodname~on the algorithm performance. Fig.\ref{fig:ablation} presents the ablation results. Here, we only present partial results due to the space limitation, and full results are presented in Appendix \ref{sec:app_exp}.

\textbf{Number of clarifying questions.}
The number of clarifying questions posed to the user denoted as $M$, is an important parameter in \methodname~framework. As illustrated in Fig.\ref{expfig:diff_hint_num}, a larger $M$ generally improves the performance of \methodname. This improvement occurs because the language model can pose more questions, gathering additional pertinent information to address the user's query. 
However, it is noteworthy that a $M$ with the value of $3$ is often sufficient for achieving commendable performance. While larger $M$ contribute to superior performance, it also imposes a heavier demand on the user to provide feedback. Therefore, it is essential to balance user experience with the quality of the model's responses. Consequently, we recommend adopting a moderate value of $M$, such as 3, for practical applications of \methodname.

\textbf{Active Learning Selection Strategy.}
\methodname~employs active learning strategy to select informative questions to inquire the user. We present the experimental results of \methodname~using three distinct active learning selection strategies, as illustrated in Fig.\ref{expfig:diff_hint_num}. The random strategy is a baseline to confirm the efficacy of the active learning techniques implemented in \methodname. Overall, the similarity strategy, which prioritizes selecting clarifying questions that are semantically similar to the user's query, outperforms the other two strategies.
An interesting result is that under specific settings of $M$, the random strategy can yield answers that surpass those generated by the DG method. This can be attributed to the fact that \methodname~is capable of producing potentially informative question candidates, from which a random selection can result in good responses.

\section{Conclusion and Future Work}

\label{sec:conclusion}

In this paper, we propose \methodname, which enables LLM to actively inquire the user and deal with ambiguous user queries. \methodname~prompts LLM to generate questions regarding the user query and utilizes active learning techniques to select the most informative questions to present to the user. Comprehensive experiments demonstrate that \methodname~clearly improves LLM's grasp of user intent and leads to more helpful answers. 
Despite the superior performance, \methodname~has certain limitations.
First, \methodname~straightforwardly prompts LLM to generate a set of questions. As shown in our case study in Sec.\ref{sec:case_study}, this process may only sometimes yield sufficient and informative questions even after the active learning process. As a potential improvement, there has been a learning-based method \cite{rag_improve} that shows promising results in generating practical clarifying questions by training a query generator. 
Second, the current embedding-based method for assessing the LLM's uncertainty about a user's query assumes that the embedding model accurately captures the semantic information of the original text. An alternative could involve estimating the model's uncertainty based on the likelihood of the generated response. 
Lastly, our experiments utilize sampled questions from existing Q\&A datasets to mimic user queries, which may only partially represent actual user interactions in the real world. It would be interesting to evaluate \methodname's performance when facing more realistic user queries and interacting with humans.
We encourage future research to explore these fields and develop more effective and efficient interactive LLMs.

\clearpage

\bibliographystyle{apalike}
\bibliography{references}

\clearpage

\appendix
\label{Appendix}

\begin{center}
\centering
	\huge{\textbf{Appendix}} %
\end{center}

\section{Discussion}

\subsection{Usage of Datasets in Experiments}

In the current research, we systematically conduct experiments using a range of well-established datasets to assess the performance of
\methodname. These experiments involve creating scenarios where user queries are ambiguous by withholding supporting facts. However, these datasets may not fully represent the complexity and nuances of real-life conversational scenarios where users pose diverse and often intricate questions.
To bridge this gap and enrich the experimental outcomes, future research could incorporate questions drawn from everyday human interactions. Additionally, exploring alternative datasets encompassing more extensive chatting histories could yield more comprehensive insights.
Besides, the current experimental framework primarily utilizes GPT-4 \cite{gpt4} as a simulated conversational partner, offering responses to clarification questions. While simulating real-world scenarios by providing GPT-4 access to supporting facts, this setup may only partially capture the conversational style typical of human interactions. The initial phase of our research involved a preliminary assessment of \methodname's competency in engaging with human participants. This provided valuable insights into its interactive capabilities.
For a more robust evaluation, it would be beneficial to expand the scope of the experiments to include a more comprehensive array of human participants. Engaging these participants in conversations with \methodname, utilizing their unique inquiries and feedback, would offer a more authentic and varied perspective on the model's interactive performance and capabilities.

\subsection{Active Learning in \methodname}

Active learning \cite{AL_inform} is acclaimed for its efficacy in augmenting model performance, particularly in scenarios characterized by a scarcity of labeled data or where obtaining such data is cost-prohibitive \cite{LM_with_al}. In the specific context of our study, \methodname~leverages this approach by judiciously selecting the most informative queries for user presentation. This strategy enables rapid learning from relevant and valuable responses, refining the model's proficiency in identifying increasingly informative queries from the available candidates. The utility of active learning extends to its capacity for tailoring interactions based on user feedback, fostering a dynamic and user-focused engagement. 

Nevertheless, there are some challenges to involving active learning techniques. A primary concern is the initial selection and crafting of clarifying questions, which must be sufficiently broad to encompass the user's potential intent while remaining focused enough to steer the model toward constructive clarifications. Furthermore, there exists a challenge concerning the potential bias inherent in the queries chosen by the active learning algorithm. Such bias could inadvertently direct the model's learning trajectory in a specific, perhaps unintended direction. Addressing this issue, future enhancements might involve the development of more advanced question-generation methods. These methods should possess a deeper understanding of the nuances and context of user queries and incorporate a more diverse array of data sources to counteract bias.
Additionally, the evolution of active learning to accommodate more intricate, multi-turn interactions could improve the richness and quality of dialogues between users and the LLM. Such advancements would likely result in more sophisticated and precise responses from the LLM, thereby enhancing the overall effectiveness and user experience.

\section{Practical Algorithm of \methodname}
\label{sec:algo}
In Sec.\ref{sec:method_workflow}, the \methodname~method is delineated through an iterative framework. Nonetheless, in practical applications, this iterative process of querying the user to diminish the uncertainty associated with the LLM's understanding of the user query may be time-intensive and onerous. To address these challenges, we introduce a streamlined version of \methodname, implemented with a singular inquiry iteration. The details of this practical implementation are outlined in Algorithm \ref{AICL_algo}.

\begin{algorithm}
\caption{Practical Implementation of \methodname}
\textbf{Input:} User input $\boldsymbol{X}$, active learning selection strategy $\gS$, active inquiry threshold $\delta$, number of clarifying questions $M$
\begin{algorithmic}
\label{AICL_algo}
\STATE Sample a set of answers to user query: $\{ \gA_i=\gM(\boldsymbol{X})\}$
\STATE Calculate the variation of the answers $\operatorname{Var}(A)$ (Eq.\ref{eqvariance})
\IF{$\operatorname{Var}(A) < \delta$}
   \STATE // Low uncertainty
   \STATE Generate the answer directly: $\boldsymbol{Y} = \gM(\boldsymbol{X})$
\ELSE
{
  \STATE // Active inquiry
  \STATE Generate a set of clarifying questions $\gQ$
  \STATE Select questions from the set with active learning strategy: $\Tilde{\gQ} = \gS(\gQ)$
  \STATE Inquire the user and get the feedback $U(\Tilde{\gQ})$
  \STATE Generate the answer $\boldsymbol{Y}$
}
\ENDIF

\end{algorithmic}

\textbf{Output:} Answer to user query $\boldsymbol{Y}$

\end{algorithm}

\section{More Experiment Details}
\label{sec:app_exp_details}

This section provides more details of the experiments, including prompts, hyper-parameters, and more examples of the datasets.

\subsection{Prompts}
\label{sec:app_prompt}

\begin{figure*}
    \includegraphics[width=1.0\linewidth]{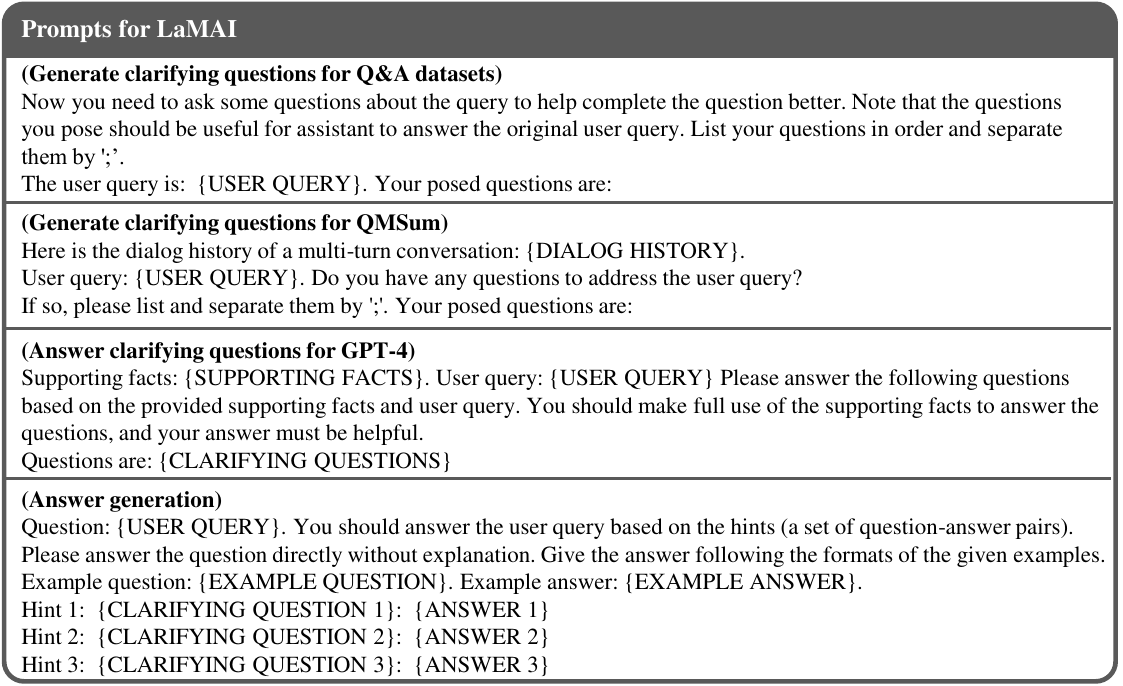}
    \caption{Prompts for \methodname.}
    \label{fig:lamai}
\end{figure*}

\begin{figure*}
    \includegraphics[width=1.0\linewidth]{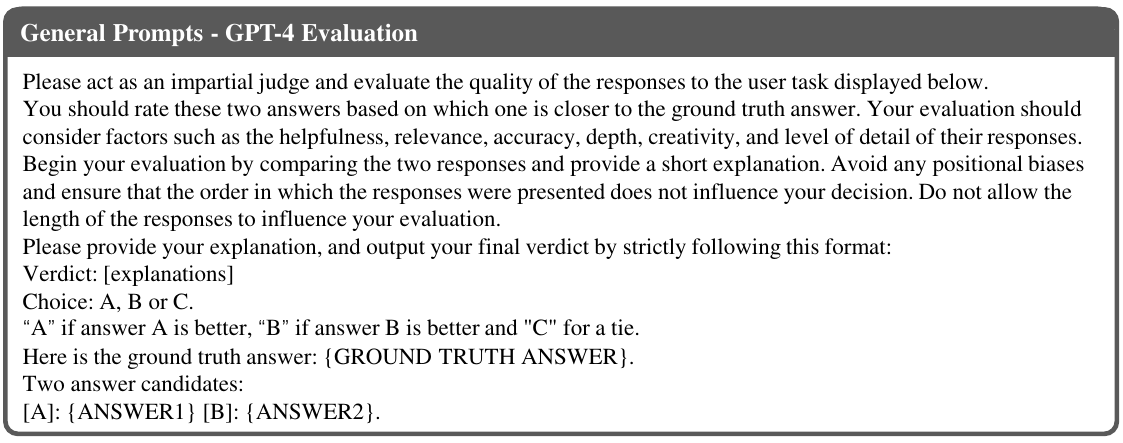}
     \caption{Prompt for GPT-4 evaluation.}
     \label{fig:eval}
\end{figure*}

\begin{figure*}
    \includegraphics[width=1.0\linewidth]{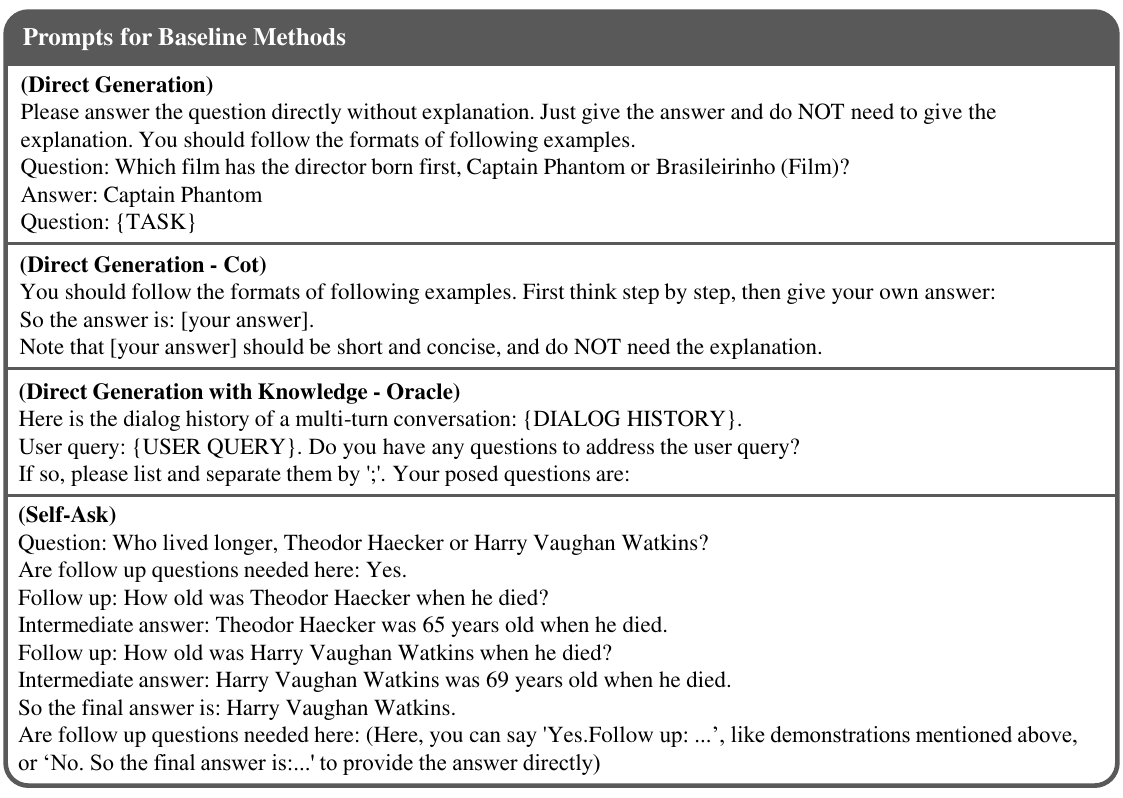}
    \caption{Prompts for baseline methods.}
    \label{fig:baseline}
\end{figure*}

We present the prompts used in the experiments in three topics: prompts for \methodname~, evaluation prompt based on \cite{zheng2023judging} and prompts for baselines, which are depicted in Fig.\ref{fig:lamai}, Fig.\ref{fig:eval}, and Fig.\ref{fig:baseline}, respectively.


\subsection{Hyper-parameters}
\label{sec:app_hyper_para}

Tab.\ref{tab:app_hyper} presents hyper-parameters used in our experiments. To implement baseline methods, we utilize their official implementation or released hyper-parameters.

\begin{table}[htbp]
    \centering
    \begin{tabular}{cc}
    \toprule
    \textbf{Name} & \textbf{Value}  \\   \toprule
    Num. of clarifying questions $M$ & 3 \\
    $\delta$ &  0.005\\
    temperature for uncertainty estimation & 0.5 \\
    top\_p & 1 \\
    presence penalty & 1 \\
    sample strategy & diversity \\
    Num. of demonstration & 2 \\
    \bottomrule
    \end{tabular}
    \caption{Hyper-parameters used in the experiments.}
    \label{tab:app_hyper}
\end{table}

\subsection{Examples of Datasets}
We present examples from datasets in Tab.\ref{tab:example_of_dataset}. The table shows that each task consists of a user query and corresponding supporting facts or user intent. \methodname~needs to propose the most valuable questions based on the context to seek clarification from the user to respond with a more refined answer.

\begin{table*}[h]
\centering
\begin{tabular}{c|p{4cm}|p{4.5cm}|p{2.5cm}}
    \toprule
    & \centering User query & Supporting facts / User intent & Label answer \\
    \midrule
    HotpotQA & Musician and satirist Allie Goertz wrote a song about the "The Simpsons" character Milhouse, who Matt Groening named after who? & "Lisa Marie Simpson is a fictional character in the animated television series, The Simpsons. She is the middle child and most intelligent of the Simpson family"... & President Richard Nixon \\ 
    \midrule
    StrategyQA & Could Lil Wayne legally operate a vehicle on his own at the beginning of his career? & Lil Wayne's career began in 1995, at the age of 12, when he was signed by Birdman and joined Cash Money Records as the youngest member of the label... & False \\ 
    \midrule
    2WikiMultiHopQA & Are director of film Move (1970 Film) and director of film Méditerranée (1963 Film) from the same country? & Move is a 1970 American comedy film... and directed by Stuart Rosenberg. The screenplay was written by... & No \\ 
    \midrule
    Musique & What is the highest point in the country where Bugabula is found? & Bugabula is one of the five traditional... It is located in the Kamuli District. Iran consists of the Iranian Plateau... & 1400 meters \\
    \midrule
    IIRC & How old was Hokutoumi when he defeated Jingaku Takashi by making him stumble out of the dohyo? & He came from the same area of Japan as future stable-mates Sakahoko and Terao. He was fond of kendo at school. He joined Izutsu stable in 1977... & 27 years \\ 
    \bottomrule
\end{tabular}

    \caption{Examples of user query, supporting facts and correct answer for the datasets used in the experiments. }
    \label{tab:example_of_dataset}
\end{table*}

\section{Additional Experiment Results}
\label{sec:app_exp}

\subsection{Full Results on QMSum}
\label{sec:app_exp_qmsum}

\begin{figure*}[ht]
    \centering
    \subfigure[Results on ES]{
    \label{fig:ablation_ES}
        \includegraphics[width=0.31\linewidth]{./figs/version1/ES.pdf}
    }
    \subfigure[Results on IS]{
    \label{fig:ablation_IS}
        \includegraphics[width=0.31\linewidth]{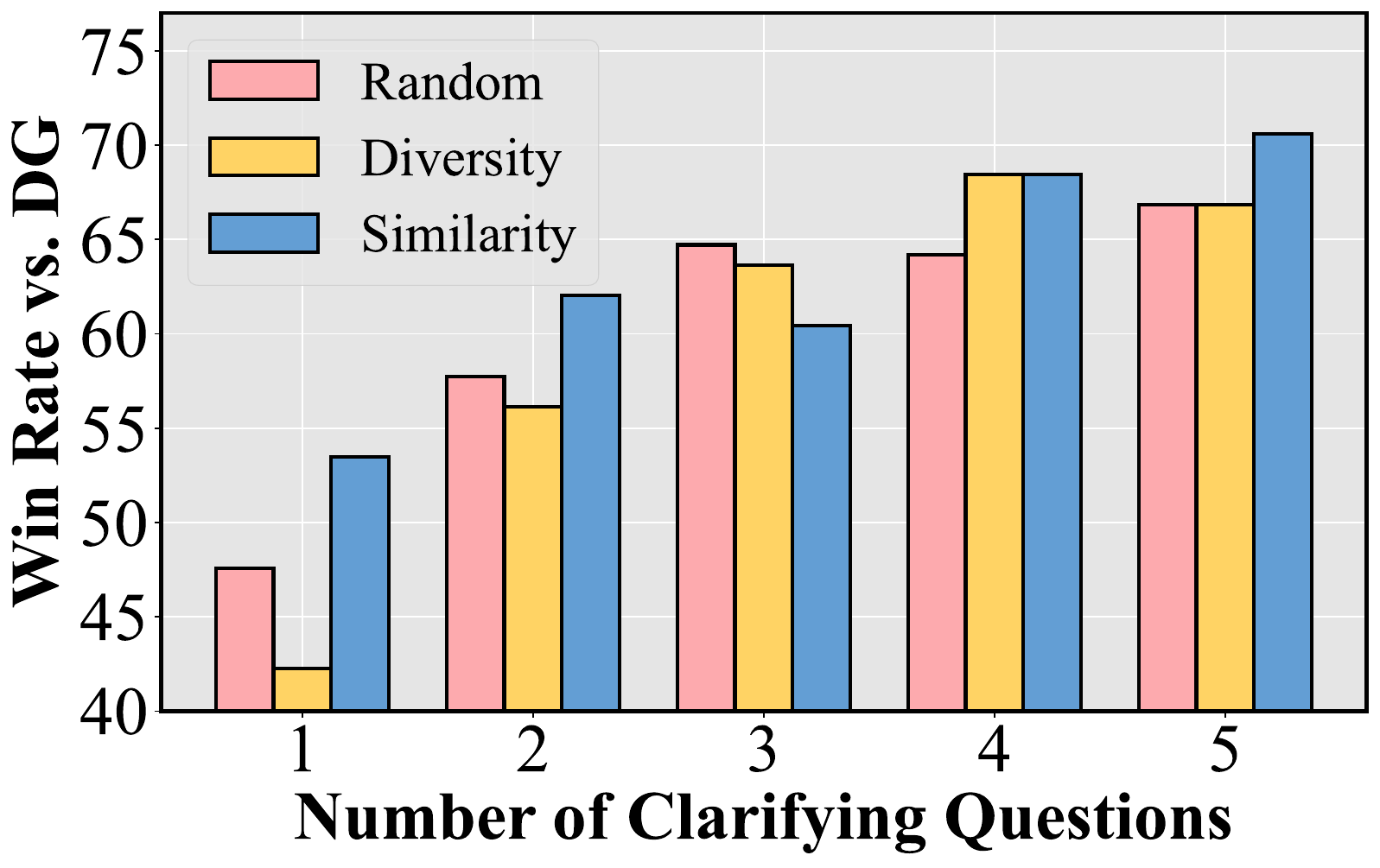}
    }
    \subfigure[Results on TS]{
    \label{fig:ablation_TS}
        \includegraphics[width=0.31\linewidth]{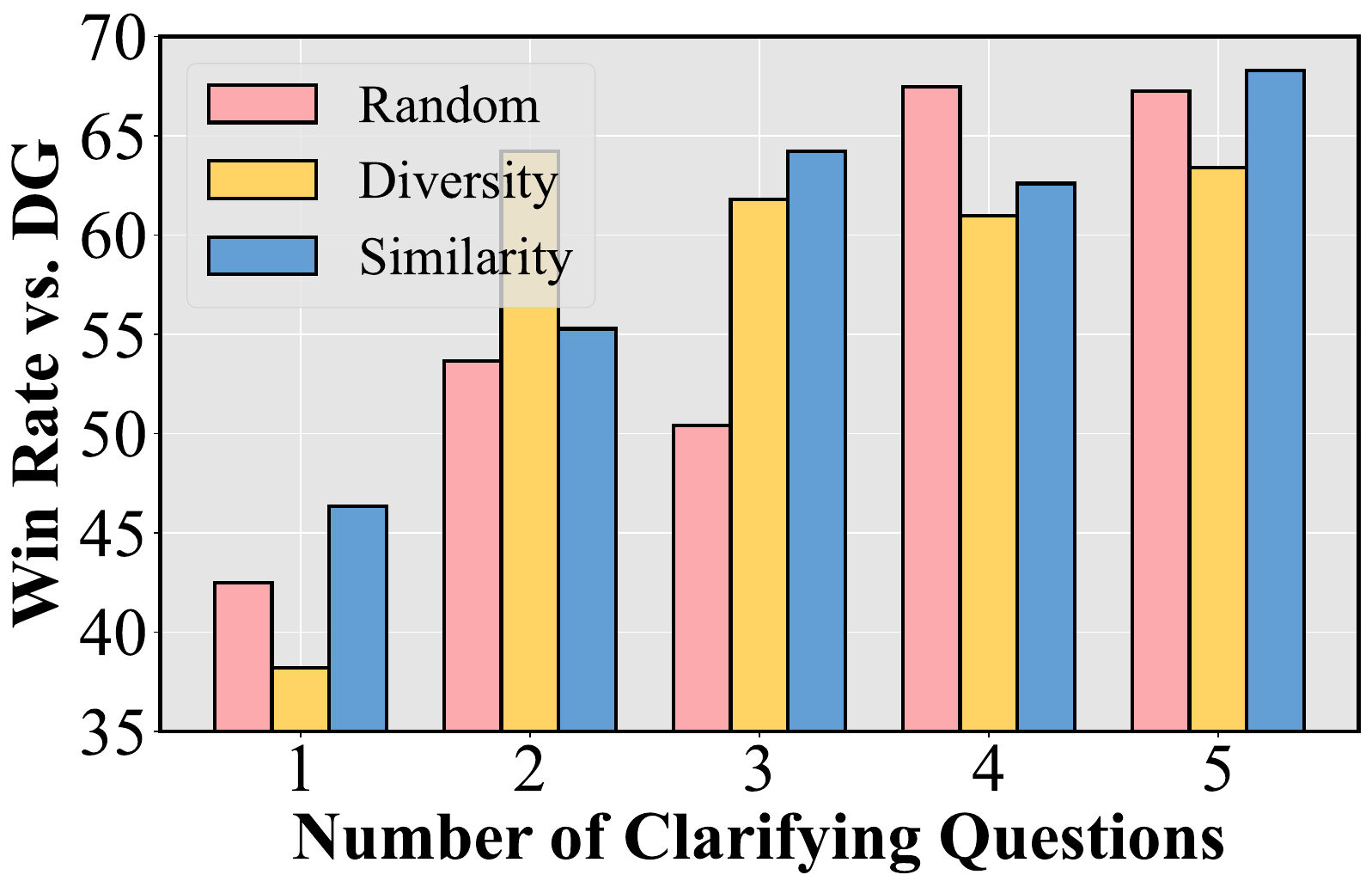}
    }
    \caption{
    Ablation study on the impact of the number of clarifying questions and active learning strategies on (a) ES dataset, (b) IS dataset, and (c) TS dataset. The values in the figure represent the win rate of \methodname~against DG method.
    }
    \label{fig:strategy}
    \end{figure*}

We present full experimental results on QMSum in Fig.\ref{fig:strategy}. Notably, there is a clear improvement in \methodname's performance correlating with an increase in the number of clarifying questions. This trend substantiates the efficacy of \methodname's clarifying questions in mitigating the contextual discrepancies between LLM and the user. Furthermore, it is observed that when the number of clarifying questions reaches a sufficiently high threshold, the performance disparities among the random, diversity, and similarity sampling strategies begin to converge. This phenomenon can be attributed to the fact that, beyond a certain point, the number of clarifying questions compensates for the need for strategic selection, rendering even random selection effective in identifying informative questions.

\subsection{Ablation Study on Active Inquiry Threshold}
\label{sec:app_exp_ablation}

In \methodname, the significance of the active inquiry threshold, denoted as $\delta$, is pivotal in determining the instances when the LLM actively seeks clarification from the user. To elucidate the impact of $\delta$ on the efficacy of \methodname, we conduct ablation experiments with different values of $\delta$. The results, as illustrated in Tab.\ref{tab:std}, indicate an enhanced performance of \methodname~at lower $\delta$ values, which correlates with a higher propensity for actively inquiring users. Notably, at a more significant threshold, e.g., $\delta=0.015$, the performance of \methodname~approximates that of the DG method, attributed to the reduced frequency of active inquiries. This observation advocates for setting a relatively low $\delta$ value to foster more proactive inquiring users by the LLM.

\begin{table*}[htbp]
\centering
\begin{tabular}{@{}c|ccc|ccc|ccc@{}}
\toprule
\multirow{2}{*}{$\delta$} & \multicolumn{3}{c|}{\textbf{ES}}  & \multicolumn{3}{c|}{\textbf{IS}}              & \multicolumn{3}{c}{\textbf{TS}}                 \\ \cmidrule(l){2-10} 
                           & \textbf{Win}                       & \textbf{Lose}             & \textbf{Tie}               & \textbf{Win}   & \textbf{Lose} & \textbf{Tie} & \textbf{Win}     & \textbf{Lose} & \textbf{Tie} \\ \midrule
0.005                        & \textbf{49.42}                     & 15.56                     & 35.02                      & \textbf{42.78} & 16.57       & 40.65      & \textbf{40.65} & 12.20       & 47.15      \\
0.010                        & \textbf{23.57}                     & 10.72                     & 65.71                      & \textbf{20.32} & 6.95       & 72.73      & \textbf{19.51} & 5.69       & 74.80      \\
0.015                        & \textbf{11.28}                     & 2.33                     & 86.39                      & \textbf{10.16} & 3.74       & 86.10       & \textbf{8.94} & 4.87       & 86.19      \\  \bottomrule
\end{tabular}
\caption{Ablation study on the effect of the active inquiry threshold on the QMSum dataset. We use GPT-4 with evaluation prompt (Fig.\ref{fig:eval}) to assess the response quality of \methodname~against DG.}
\label{tab:std}
\end{table*}

\subsection{Performance on User Query with Less Context}

\methodname~actively seeks clarification from users to gain additional insights regarding their queries. To evaluate this capability for clarification-seeking, experiments were conducted utilizing the QMSum dataset, with varying masking applied to the dialogue history. As the masking rate increases, the availability of contextual information to the LLM correspondingly diminishes. As the experimental results are shown in Tab.\ref{tab:mask}, as more significant proportions of contextual data are obscured, \methodname~increasingly outperforms the DG method. This enhanced performance is attributed to \methodname's proactive clarification-seeking strategy, which mitigates the challenges of insufficient contextual data. Furthermore, it is observed that \methodname~maintains its efficacy in eliciting additional information helpful for answering the user query, even when the whole context is available, thereby enriching its responses to user inquiries.

\begin{table*}[htbp]
\centering
\begin{tabular}{@{}c|ccc|ccc|ccc@{}}
\toprule
\multirow{2}{*}{Mask Rate} & \multicolumn{3}{c|}{\textbf{ES}}  & \multicolumn{3}{c|}{\textbf{IS}}              & \multicolumn{3}{c}{\textbf{TS}}                 \\ \cmidrule(l){2-10} 
                           & \textbf{Win}                       & \textbf{Lose}             & \textbf{Tie}               & \textbf{Win}   & \textbf{Lose} & \textbf{Tie} & \textbf{Win}     & \textbf{Lose} & \textbf{Tie} \\ \midrule
0                          & \multicolumn{1}{l}{\textbf{47.86}} & \multicolumn{1}{l}{34.24} & \multicolumn{1}{l|}{17.90} & \textbf{48.13} & 36.90         & 14.97        & \textbf{55.28}   & 30.08         & 14.63        \\
0.3                        & \textbf{55.14}                     & 23.67                     & 21.22                      & \textbf{58.70} & 31.52       & 9.78       & \textbf{48.96} & 34.38       & 16.67      \\ 
0.5                        & \textbf{63.04}                     & 23.35                     & 13.62                      & \textbf{63.64} & 21.93       & 14.44      & \textbf{61.79} & 20.33       & 17.89      \\
0.7                        & \textbf{66.17}                     & 21.05                     & 20.54                      & \textbf{71.20} & 13.91       & 15.71      & \textbf{62.12} & 25.00       & 12.88      \\ \bottomrule
\end{tabular}
\caption{Performance of \methodname~on user query with different levels of context masking. We use GPT-4 with evaluation prompt from Fig.\ref{fig:eval} to assess the response quality of \methodname~against DG. \methodname~consistently surpasses DG under different levels of context masking. }
\label{tab:mask}
\end{table*}

\subsection{More Algorithm Running Examples}

\begin{figure*}[ht]
    \centering
    \includegraphics[width=0.95\linewidth]{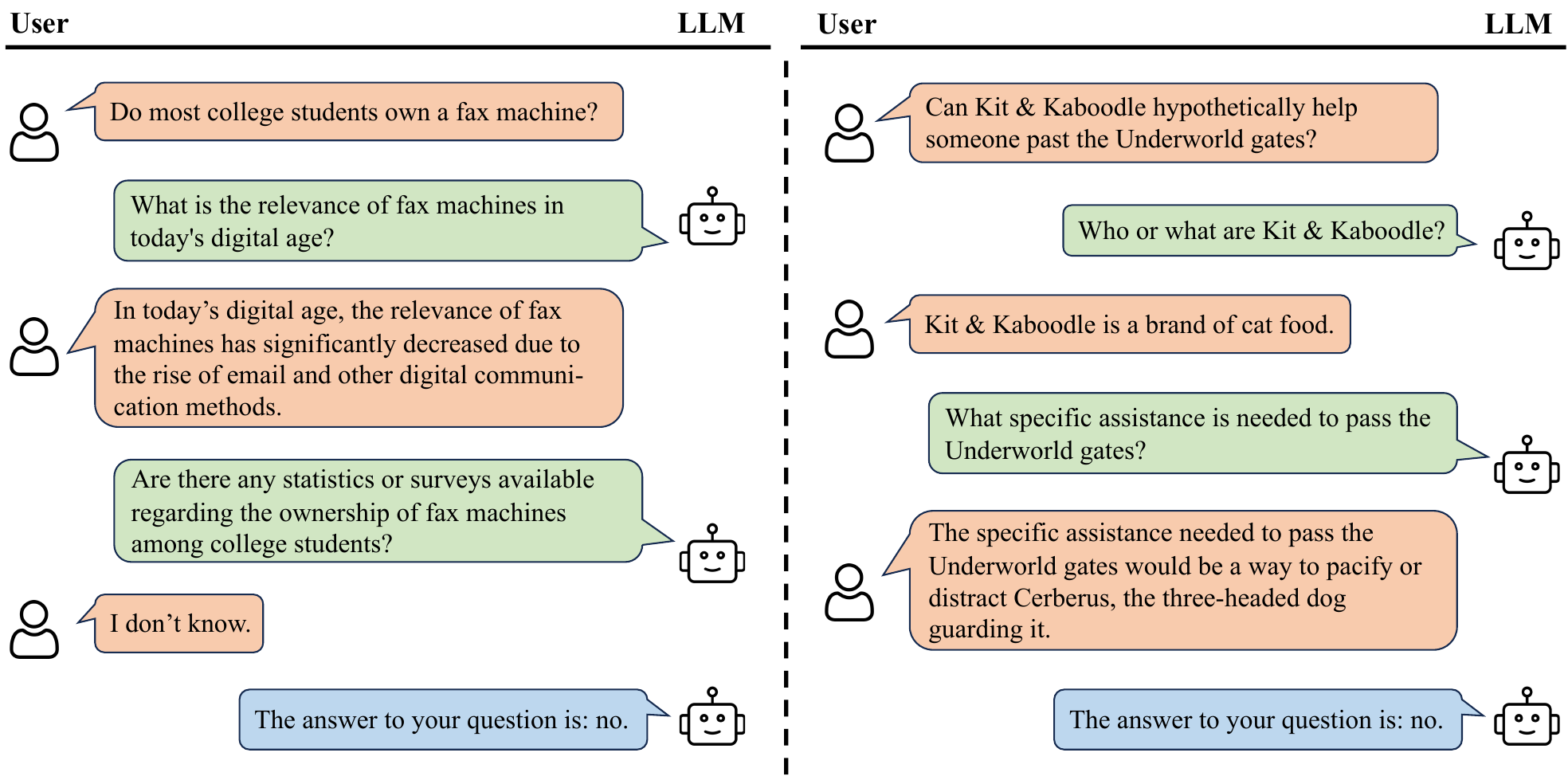}
    \caption{
        Examples of model output for \methodname. The green chat frame presents the clarifying questions that LLM actively inquires the user.
    }
    \label{fig:example_of_model_output_app}
\end{figure*}

\begin{figure*}[ht]
    \centering
    \includegraphics[width=0.95\linewidth]{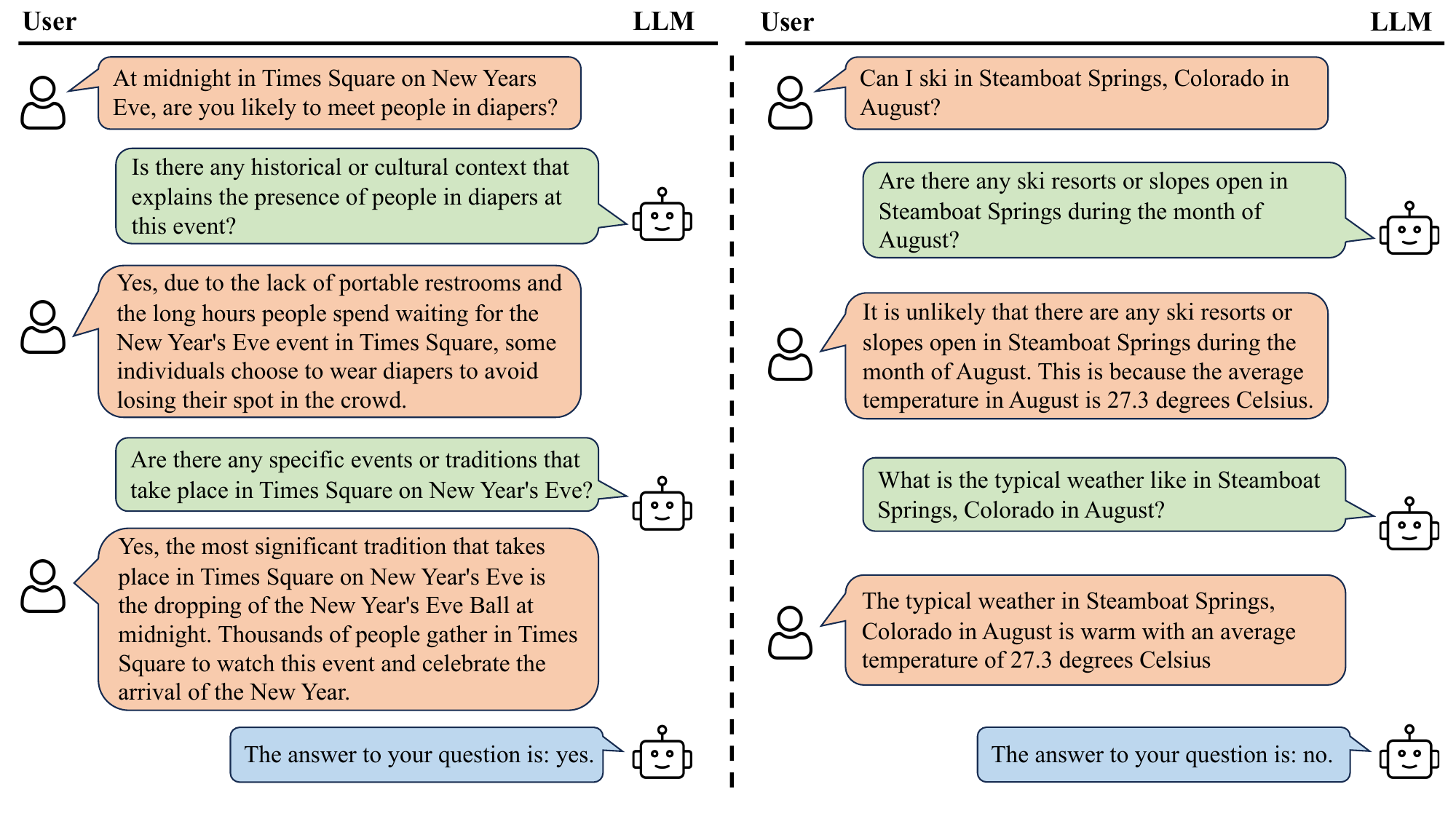}
    \caption{
        Examples of model output for \methodname. The green chat frame presents the clarifying questions that LLM actively inquires the user.
    }
    \label{fig:example_of_model_output_app_2}
\end{figure*}

\begin{figure*}[ht]
    \centering
    \includegraphics[width=0.95\linewidth]{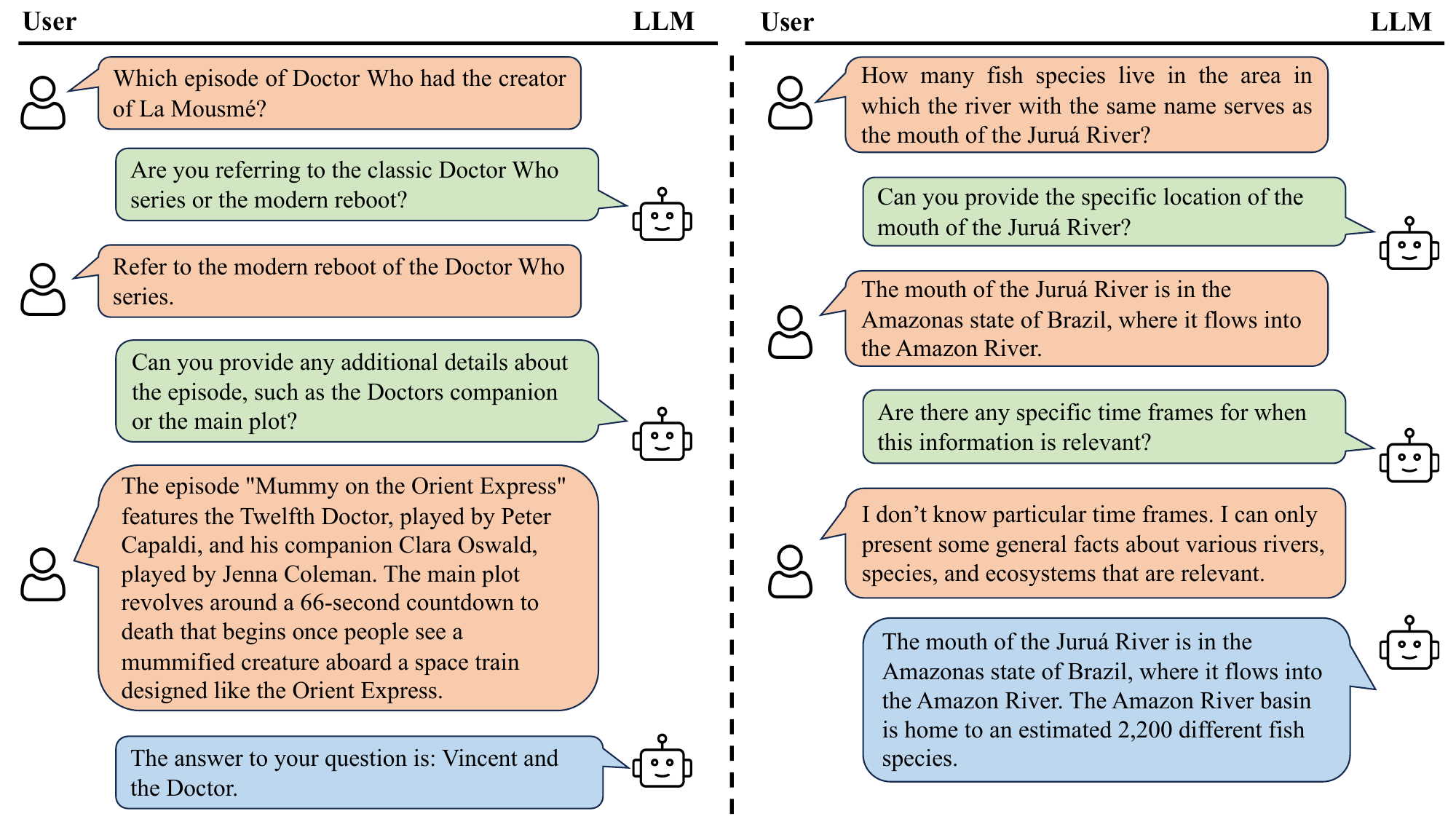}
    \caption{
        Examples of model output for \methodname. The green chat frame presents the clarifying questions that LLM actively inquires the user.
    }
    \label{fig:example_of_model_output_app_3}
\end{figure*}

\begin{figure*}[ht]
    \centering
    \includegraphics[width=0.95\linewidth]{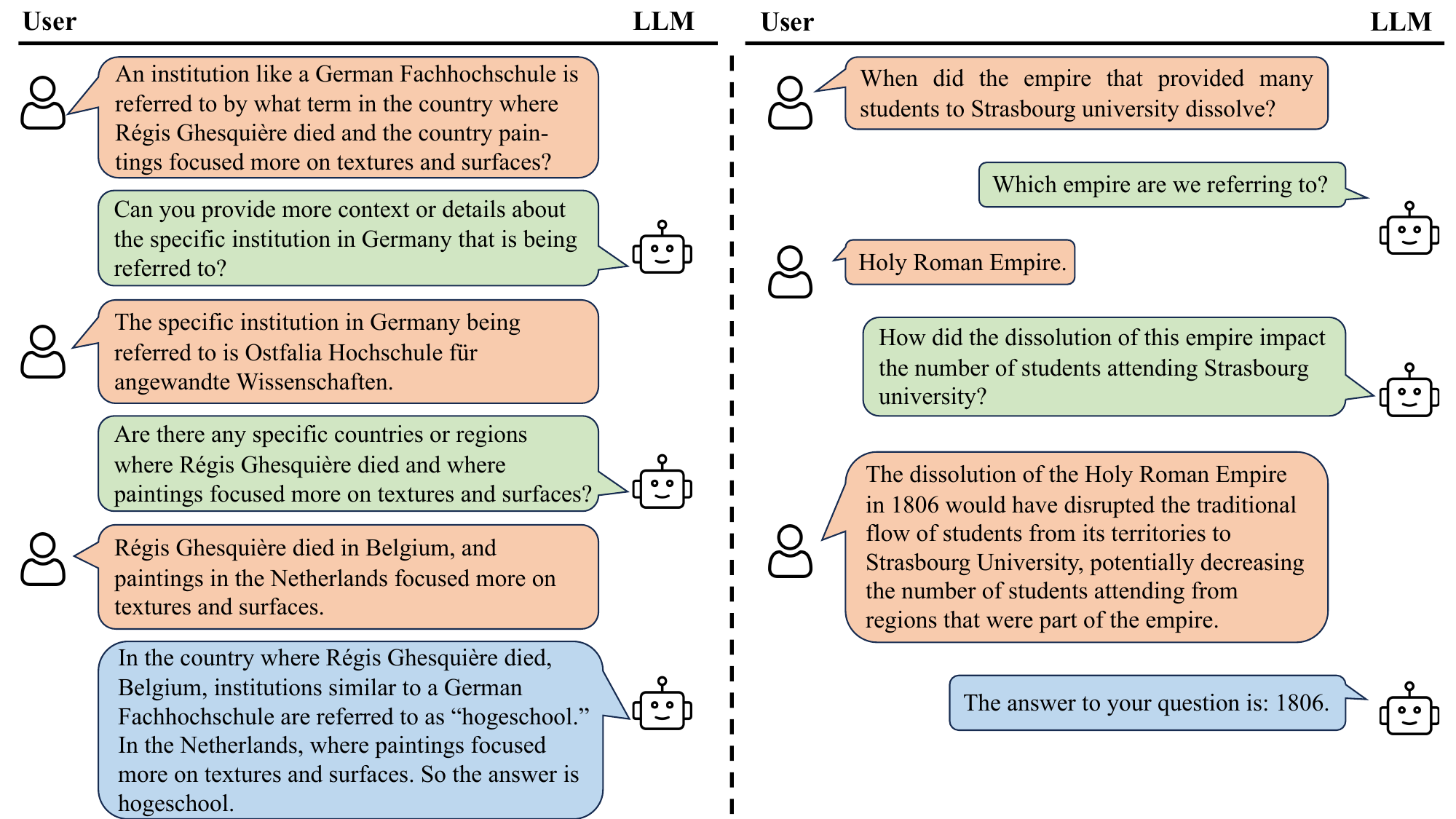}
    \caption{
        Examples of model output for \methodname. The green chat frame presents the clarifying questions that LLM actively inquires the user.
    }
    \label{fig:example_of_model_output_app_4}
\end{figure*}

We provide more examples of LLM output during \methodname~running process in Fig.\ref{fig:example_of_model_output_app}, Fig.\ref{fig:example_of_model_output_app_2}, Fig.\ref{fig:example_of_model_output_app_3} and Fig.\ref{fig:example_of_model_output_app_4}. It is evident that \methodname~poses useful clarifying questions regarding the ambiguous user query, thereby enhancing the interpretation of user intent.

\clearpage

\section{Retrievals for Notations and Abbreviations}

Tab.\ref{tab:notations} presents a list of the notations and abbreviations employed throughout this paper, serving as a convenient reference for the reader.

\begin{table*}[h]
    \centering
    \begin{tabular}{cc}
    \toprule
    \textbf{Name} & \textbf{Meaning}  \\   \toprule
    \textbf{Notations} \\ 
    $\delta$ & threshold for active inquiry \\
    $\gM$ & large language model \\
    $\boldsymbol{X}$ & user query \\
    $\boldsymbol{Y}$ & LLM's answer \\
    $\gS$ & sampling strategy of active learning \\
    $\gQ$ & clarifying questions (before selection)  \\
    $\Tilde{\gQ}$ & clarifying questions (after selection)  \\
    $M$ & number of clarifying questions  \\
    $E_i$ & embedding of the text \\
    \toprule
    \textbf{Abbreviations} \\
    \methodname & Language Model Active Inquiry \\
    LLM & Large Language Model \\ 
    DG & Direct Generation \\ 
    CoT & Chain-of-Thought \\ 
    RAG & Retrieval-Augmented Generation \\ 
    AIM & Active Inquiry Module \\ 
    Q\&A & Question-Answering \\ 
    \bottomrule
    \end{tabular}
    \caption{Notations and abbreviations in this paper.}
    \label{tab:notations}
\end{table*}

\end{document}